\documentclass[acmlarge]{acmart}
\AtBeginDocument{%
  \providecommand\BibTeX{{%
    \normalfont B\kern-0.5em{\scshape i\kern-0.25em b}\kern-0.8em\TeX}}}

\usepackage{algorithm}
\usepackage{algpseudocode} 
\usepackage{multirow}
\usepackage{graphics}
\usepackage{amsmath}
\usepackage{upgreek}
\usepackage{subcaption}

\DeclareMathOperator*{\argmax}{argmax}

\begin{document}

\title{Masked Deep Q-Recommender for Effective Question Scheduling}

\author{Keunhyung Chung}
\authornote{Both authors contributed equally to this research.}
\email{keunhyung_chung@tmax.co.kr}
\author{Daehan Kim}
\authornotemark[1]
\email{daehan_kim@tmax.co.kr}
\author{Sangheon Lee}
\email{sangheon_lee@tmax.co.kr}
\author{Guik Jung}
\email{guik_jung@tmax.co.kr}
\affiliation{%
  \institution{TmaxEdu Inc.}
  \city{Seongnam}
  \state{Gyeonggi}
  \country{South Korea}
}

\renewcommand{\shortauthors}{Anonymous Author(s)}

\begin{abstract}
 Providing appropriate questions according to a student's knowledge level is imperative in personalized learning. However, It requires a lot of manual effort for teachers to understand students' knowledge status and provide optimal questions accordingly. To address this problem, we introduce a question scheduling model that can effectively boost student knowledge level using Reinforcement Learning (RL). Our proposed method first evaluates students' concept-level knowledge using knowledge tracing (KT) model. Given predicted student knowledge, RL-based recommender predicts the benefits of each question. With curriculum range restriction and duplicate penalty, the recommender selects questions sequentially until it reaches the predefined number of questions. In an experimental setting using a student simulator,  which gives 20 questions  per day  for two weeks, questions recommended by the proposed method increased average student knowledge level by 21.3\%, superior to an expert-designed schedule baseline with a 10\% increase in student knowledge levels.
\end{abstract}




\maketitle

\section{Introduction}

Many online learning platforms like Coursera and Udacity are getting more and more attention in this era of distant learning. Unlike face-to-face learning environments, online learning has limited clues to grasp student knowledge status, since video lectures, question sheets, or textbooks are one-way material where interactions are scarce. To facilitate effective online learning, two main questions need to be solved: 
\begin{itemize}
    \item How can we quantitatively measure student performance on certain questions or concepts?
    \item How can we design a personal learning schedule in a way that maximizes his/her performance within a given period?
\end{itemize}

We focused on the second problem of \textbf{providing an adequate sequence of questions in a personalized way}. In this work, we introduce MDQR (Masked Deep-Q Recommender) that can effectively boost student performance considering students' current curriculum and their knowledge level. Our proposed model architecture is composed of two parts: student simulator and recommender. Since we cannot feed problems to real students, we need knowledge tracing (KT) model, which can approximate students' probability of answering the specific questions correctly. Whenever the recommender gives a sequence of questions, they are feed to the student simulator, and responses are sampled from the simulator's predicted probabilities - either correct or wrong. The recommender is then reinforced by reward which consists of a change in student performance. Here, the student performance can be defined as the average correct probability over the student's test range.

Experimental results show that the proposed method improves \textbf{11.2\%} more knowledge level increase over expert-selected recommendation policy in average test range when evaluated in a math question dataset collected AIHUBmath.

\section{Backgrounds}

\subsection{Reinforcement Learning}
RL is a branch of machine learning that is used to optimize a continuous decision problem. In RL, the agent, the acting subject, repeatedly interacts with the environment. At each time-step $t$, the agent observes the state of the environment, which is described as $s_t$, and the agent determines action $a_t$ according to the state $s_t$. The mapping function that receives the state as input and decides the action is called a policy $\pi$, and it can be described as $a_t = \pi(s_t)$. After the agent selects action $a_t$, the environment takes the state-action pair $(s_t, a_t)$ and transits to the next state $s_{t+1}$. The environment also emits a reward $r_t$ for each instantaneous transition $(s_t, a_t, s_{t+1})$. In the next type-step $t+1$, the agent receives the next state $s_{t+1}$ and decides the next action $a_{t+1}$ by the policy $\pi$, and this process continues until the end of an episode. The goal of RL is to find a policy $\pi$ that maximizes the discounted sum of future rewards received from the environment. This discounted sum of future rewards are called as \textit{return} and \textit{return} at time $t$ is defined as $R_t = \sum_{t'=t}^{T} \gamma^{t'=t}r_{t'}$, where $\gamma$ is a discount factor and $T$ is the time-step at which the episode ends.

\textit{Q-learning} \cite{watkins1992q} is one of the popular algorithms for finding a policy that maximizes \textit{return}. \textit{Q-learning} uses the optimal action-value function $Q^*(s, a)$ which is the expected \textit{return} achievable by following the best policy, after seeing some state $s$ and then taking some action $a$, $Q^*(s, a) = \max_{\pi}\mathbb{E}[R_t|s_t=s, a_t=a, \pi]$. This algorithm find $Q^*(s, a)$ by updating an action-value function (which is also called as \textit{Q-function}) repeatedly using an important theory know as the \textit{Bellman equation}, 
\begin{equation}
    Q^*(s, a) = \mathbb{E}_{r, s'}[r + \gamma \max_{a'} Q^*(s', a')|s, a]
\end{equation}
Using iteration $Q_{i+1}(s, a) \leftarrow r + \gamma \max_{a'} Q_i(s', a')$, a \textit{Q-function} can converges to the optimal \textit{Q-function} ($Q_i(s, a) \rightarrow Q^*(s, a), \: \forall (s, a)$ as $i \rightarrow \infty$)\cite{sutton2018reinforcement} and $r + \gamma \max_{a'} Q_i(s', a')$ is called the target value.

In practice, we approximate \textit{Q-function} because it is impossible to represent all possible $Q(s, a)$ separately if there are a huge number of state-action pairs $(s, a)$. There have been many follow-up studies that approximate \textit{Q-function}, and DQN \cite{mnih2013playing, mnih2015human} is the first to use convolutional neural networks to approximate \textit{Q-function}. DQN trains an agent to play Atari 2600 games \cite{bellemare2013arcade} and use \textit{Q-function} $Q_\theta(s, a)$ where $s$ is a set of raw pixel images from Atari 2600 games and $\theta$ are parameters of neural networks. This algorithm use replay memory \cite{lin1992reinforcement} which saves the transition $(s_t, a_t, r_t, s_{t+1})$ when an agent plays with the environment and randomly samples previous transitions when the agent is trained. Another characteristic of DQN is that it uses another Q-network called target network for the stability of training. The target network is used to calculate the target value $r + \gamma \max_{a'} Q_{\theta'}(s', a')$ where $\theta'$ are parameters of the target network and $\theta'$ are only updated with original Q-network parameters $\theta$ for every some intervals. In this paper, we used a variant of DQN algorithm to schedule the questions efficiently.

\subsection{Knowledge Tracing}
KT is a task that evaluates a student's knowledge level based on log data accumulated while learning. Student log data has various types of features such as timestamp, student action type, and time spent on action. Among them, question information and correct answers are mainly used for the KT task. The KT model infers the student's mastery levels one each knowledge concept based on the student's past question-solving log data. The output of the model is the probability that the student will correct the question corresponding to each concept, which can be interpreted as the student's mastery of the corresponding knowledge concept. For the concept $c_t$ of the question provided to the student at time $t$, the correct answer predicted by the model as $p_t$, and the feedback result from the student as $f_t$ , the model is trained through the following binary cross-entropy loss.

\begin{equation}
    L = -\sum_t (f_t \log{p_t} + (1-f_t) \log{(1-p_t)})
\end{equation}

The first proposed deep learning-based KT technique is the Deep Knowledge Tracing (DKT) \cite{piech2015dkt}. They used a recurrent neural network (RNN) \cite{williams1989rnn} that mainly deals with time-series data, focusing on the fact that the log data of the student's question solving is sequential data. For the input vectors $x_1, ..., x_t$ of the model, the student's knowledge concept mastery probability $y_1, ..., y_t$ is calculated as follows.

\begin{align}
    y_t &= sigmoid ( h_t \cdot W_{yh} + b_y )    \\
    h_t &= tanh ( x_t \cdot W_{hx} + h_{t-1} \cdot W_{hh} + b_h )
\end{align}
where $h_t$ is the hidden state vector of the RNN, which is information that implies the student's question-solving results up to time t. The parameters of the model are composed of an input weight matrix $W_{hx}$, a hidden state weight matrix $W_{hh}$, an output weight matrix $W_{yh}$, and biases, which are updated through training. After the DKT model was proposed, more advanced deep learning models than RNN dealing with sequential data were applied to the KT task.

In this paper, we used the Bi-LSTM-based (NPA) model with self-attention \cite{lee2019creating} as a student simulator. Unlike the original NPA model, we use the difficulty of question along with the knowledge concept of the question and the results of student feedback. The difficulty of the question affects the change in the student's knowledge level \cite{ma2016effect}. Therefore, the KT model using this information can expect more precise knowledge level prediction. Also, using the student's knowledge level based on the difficulty of the question, our recommender can schedule questions by specifying the difficulty of the question.

\begin{figure}[h]
  \centering
  \includegraphics[width=0.7\linewidth]{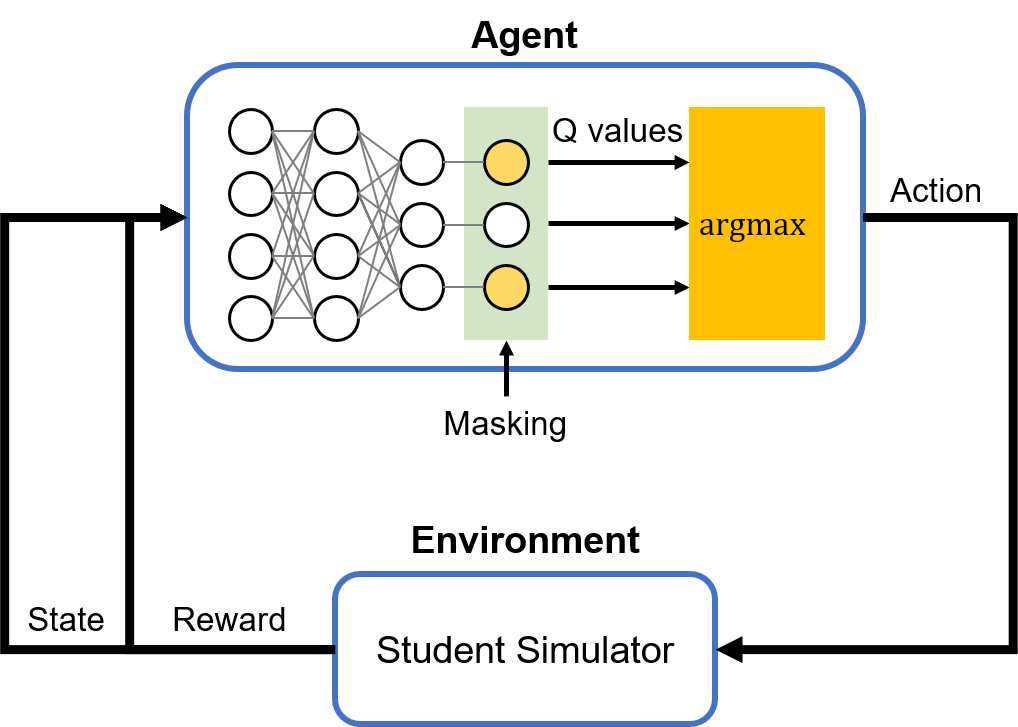}
  \caption{Structure of MDQR and its simulation flow as Markov Decision Process}
  \Description{Structure of MDQR.}
  \label{fig:mdqr-structure}
\end{figure}

\begin{figure}[h]
  \centering
  \includegraphics[width=0.75\linewidth]{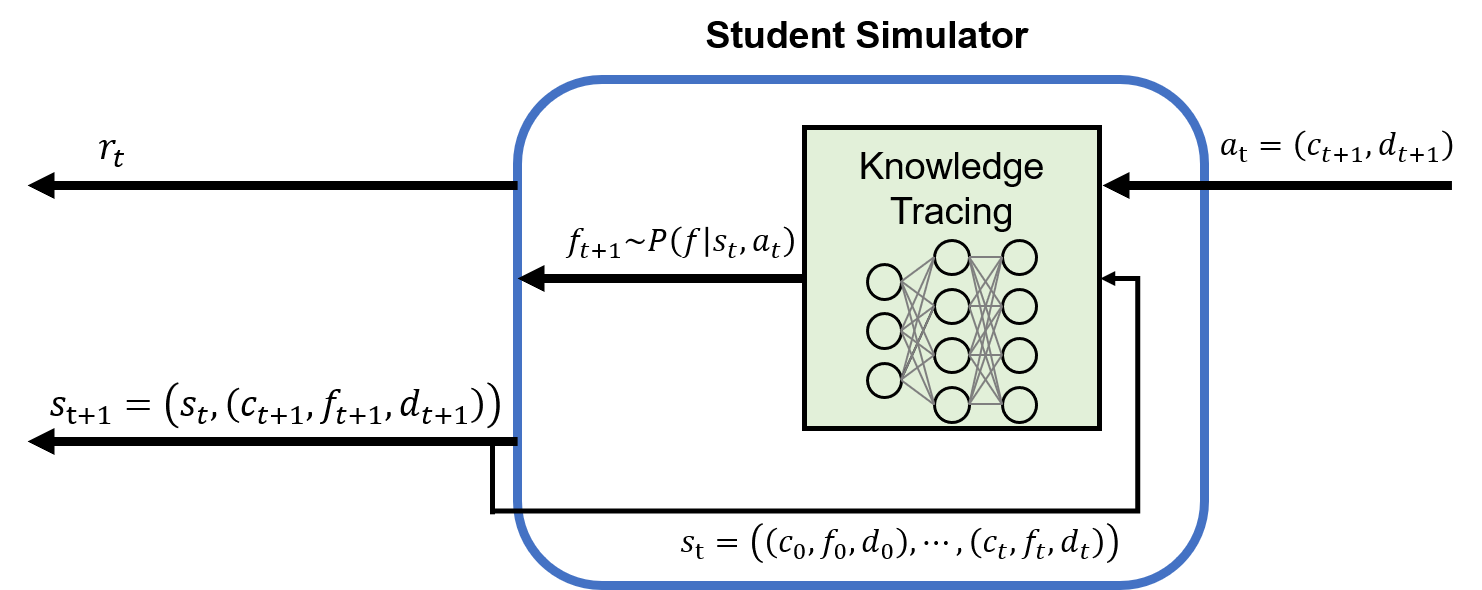}
  \caption{The student simulator works based on the Knowledge Tracing model.}
  \Description{The structure of the student simulator}
  \label{fig:student-simulator}
\end{figure}

\section{Problem Formulation}



The problem we are trying to solve is to schedule a curriculum for a specific range of questions so that a student's knowledge level will increase as much as possible within a limited time. To solve this problem, we first constructed a scenario in which students were recommended 20 questions a day for two weeks, where the recommended questions are limited to a test range $C$ (e.g., the final exam in the second semester of grade 8). A question is expressed as $(c, d)$, which is a pair of knowledge concept $c$ and the difficulty of the question $d$. The student's knowledge level on each $(c, d)$ is measurable by the KT model and the goal of our scheduler is to find the best sequence of questions that can maximize 
\begin{equation}
    \frac{1}{|C|} \sum_{(c, d) \in C}l_{c, d},    
\end{equation}
where $l_{c, d}$ is a knowledge level on $(c, d)$ measured by the KT model. We also assumed that the student's knowledge level changes once a day for the simplicity of the scheduling simulation.



\subsection{Markov Decision Process}

To apply RL framework to the question scheduling simulation, we model the scheduling scenario as a Markov Decision Process (Figure \ref{fig:mdqr-structure}).

\subsubsection{State}
The state is the student's question solving history and it can be composed of knowledge concept of the question $c$, a student's  feedback result $f$ (i.e., correct or  wrong), and question difficulty $d$. We denote $(c, f, d)$ as a \textit{question-feedback} and state at time $t$ is a sequence of \textit{question-feedback}s, i.e.,
\begin{equation}
  s_t = ((c_0, f_0, d_0), \cdots, (c_t, f_t, d_t)).
\end{equation}
the difficulty $d$ has three levels which are easy, medium, hard.

\subsubsection{Action}
At each time $t$, MDQR can select which knowledge concept of the question and at what difficulty level of question to assign to the student, i.e., 
\begin{equation}
    a_t = (c_{t+1}, d_{t+1}).
\end{equation}

\subsubsection{State Transition}
After the student solve the question which is recommended by action $a_t = (c_{t+1}, d_{t+1})$, the student's feedback result $f_{t+1}$ is sampled by probability distribution $P(f|s_t, a_t)$. The probability $P(f|s_t, a_t)$ is a prediction of whether the feedback result $f_{t+1}$ will be correct when the current state is $s_t$ and action (a recommended question) is $a_t$. We modeled The function $P$ by KT (Figure \ref{fig:student-simulator}).

\begin{algorithm}[H]
    \caption{Masked Deep Q-Recommender}
    \begin{algorithmic}[1]
        \For{episode $= 1$ to $M$}
            \State Initialize state $s_1$
            \For{$t = 1$ to $T$}
                \For{$e = e_1$ to $e_N$}
                    \State Get a random action $a_t \in C_\mathrm{e}$ with probability $\epsilon$
                    \State otherwise $a_t \gets \argmax_{a \in C_\mathrm{e}} Q_\theta(s_t, a)$
                    \State Execute action $a_t$ and observe reward $r_t$ and next state $s_{t+1}$ from the student simulator
                    \State Store transition $(s_t, a_t, r_t, s_{t+1})$ in replay memory $D_\mathrm{e}$
                \EndFor
                \For{$e = e_1$ to $e_N$}
                    \State Sample random minibatch of transitions $\mathrm{trans}_j = (s_j, a_j, r_j, s_{j+1})$ from $D_\mathrm{e}$
                    \State Update Q network with a gradient descent step on DQNLoss$(\mathrm{trans}_j, Q_\theta, Q_{\theta'}, C_\mathrm{e})$:
                    \begin{displaymath}
                        \theta \gets \theta - \alpha\nabla_\theta\sum_j\text{DQNLoss}(\mathrm{trans}_j, Q_\theta, Q_{\theta'})
                    \end{displaymath}
                \EndFor
                \State Update the target network:
                \begin{displaymath}
                    \theta' \gets \tau\theta + (1-\tau)\theta'
                \end{displaymath}
            \EndFor
        \EndFor
        \State
        \Function{DQNLoss}{$\mathrm{trans}_j, Q_\theta, Q_{\theta'}, C_\mathrm{e}$}
            \State $y_j =
                \begin{cases}
                            r_j,& \text{\small if episode terminates at step $j+1$}\\
                            r_j + \gamma\max_{a' \in C_\mathrm{e}}Q_{\theta'}(s_{j+1}, a'),& \text{\small otherwise}
                \end{cases}$
            \State \Return $(y_j - Q_\theta(s_j, a_j))^2$
        \EndFunction
    \end{algorithmic}
    \label{algorithm:mdqr}
\end{algorithm}

\subsubsection{Reward}
The purpose of our recommendation model is to raise the student's knowledge level above the initial state. We assumed that the probability of correcting a question is the student's level of knowledge. Therefore, the reward at time $t$ is calculated using the correctness probability function $P$ as follows:

\begin{equation} \label{rwd_eq}
    r_t = k_t - \psi_t,
\end{equation}
Where
\begin{equation}
    k_t = \sum_{a \in C_\mathrm{e}} (P(f = 1 | s_{t+1}, a) - P(f = 1 | s_t, a)),
\end{equation}
and
\begin{equation}
    \psi_t = 
    \begin{cases}
        \uppsi,& \text{\small if $a_t \in \{a_{t-20}, \cdots, a_{t-1}\}$} \\
        0,& \text{\small otherwise}
    \end{cases}
\end{equation}
$k_t$ represents the increments in the knowledge level at state $s_{t+1}$ compared to state $s_t$. $C_\mathrm{e}$ is a set of questions in test range $\mathrm{e}$. $\psi_t$ is a penalty term that prevents duplicate recommendations for the same question. We give a duplicate penalty $\uppsi$ if the recommended question was included in the last 20 questions.

\subsection{Masked Deep Q-Recommender}
Like DQN, MDQR has \textit{Q-function} $Q_\theta (s, a)$ to determine the action at state $s$. For MDQR agent to select a question in a test range $C_\mathrm{e}$, we mask a question outside the test range $C_\mathrm{e}$ when searching the maximum value of $Q_\theta (s, a)$, i.e., $a_{t+1} = \argmax_{a \in C_\mathrm{e}} Q_\theta(s_t, a)$.

To train the MDQR model, we also use the DQN training method. DQN collects training data through simulation and stores it in replay memory, and the data collected in this replay memory is used when training the model. MDQR also collects training data through scheduling simulations, and the simulations are performed under each test range so that one MDQR agent can cover all test ranges. This is because the MDQR agent cannot learn the scheduling of questions in a specific test range if the training data is collected under the simulation where the MDQR agent recommends questions without limiting the test range. Therefore, unlike DQN, MDQR has as many replay memories as the number of test ranges. The MDQR algorithm is illustrated in Algorithm \ref{algorithm:mdqr}

\section{Evaluation}

\subsection{Dataset}

We use the AIHUBmath dataset from \textit{National Information society Agency}. AIHUBmath is a log of solving math questions for grades 7-9. There are 707,450 question-solving interactions from 4,673 students. The average number of interactions per student is 153.84. Table \ref{tab:concepts-examples} shows examples of knowledge concepts contained within each test range in AIHUBmath.

\begin{table}[H]
  \begin{tabular}{c|l|l}
    \toprule
     Concept Id&Concept Name&Test Range \\
    \hline
    \midrule
    1 & Prime number and composite number & 7th grade 1st semester midterm \\
    144 & Quadrant & 7th grade 1st semester final \\
    218 & Conditions for congruence of triangles & 7th grade 2nd semester midterm \\
    314 & Frequency polygon & 7th grade 2nd semester final \\
    \hline
    343 & Representing repeating decimal to fraction & 8th grade 1st semester midterm \\
    461 & Finding slope of linear function graph & 8th grade 1st semester final \\
    519 & Circumcenter and incenter of triangles & 8th grade 2nd semester midterm \\
    647 & Properties of probability & 8th grade 2nd semester final \\
    \hline
    732 & Factorization formula & 9th grade 1st semester midterm \\
    790 & Graph of quadratic function & 9th grade 1st semester final \\
    813 & Trigonometric ratio & 9th grade 2nd semester midterm \\
    886 & Finding variance and standard deviation & 9th grade 2nd semester final \\
    \bottomrule
  \end{tabular}
  \caption{Examples of Knowledge Concepts}
  \label{tab:concepts-examples}
\end{table}

\subsection{Scheduling Methods}

We simulate the following three recommended methods and compare their performance: \textit{random}, \textit{expert}, MDQR. \textit{Random} is a method in which the concept and difficulty of questions are randomly asked within a specific test range. \textit{Expert} ask questions in the order of the curriculum within a specific test range. The difficulty level of the question is adaptively adjusted according to the student's previous question-solving record. If the previous question is correct, the next question is more difficult than the previous question (until it reaches the highest difficulty), and if it is wrong, it is more easy than the previous question difficulty (until it reaches the lowest difficulty). MDQR recommends a question $a$ which maximizes $Q_\theta(s, a)$ within a specific test range.

\subsection{Experimental Design}

Assuming that the question is recommended according to the current grade of the student, the scheduling simulation was also conducted so that the question was asked within a specific range. In addition, we conducted the scheduling simulation with 20 questions a day for 2 weeks.

\begin{table}[H]
  \begin{tabular}{l||c|c|c}
    \toprule
    Scheduling Method&Random&Expert&MDQR \\
    \hline
    \midrule
    7th grade 1st semester midterm & 0.1176 & 0.1416 & \textbf{0.2350} \\
    7th grade 1st semester final & 0.1544 & 0.1552 & \textbf{0.2363} \\
    7th grade 2nd semester midterm & 0.0455 & 0.0863 & \textbf{0.1502} \\
    7th grade 2nd semester final & -0.0340 & -0.1450 & \textbf{0.1358} \\
    \hline
    8th grade 1st semester midterm & 0.1295 & 0.0180 & \textbf{0.2669} \\
    8th grade 1st semester final & 0.0996 & 0.1357 & \textbf{0.1711} \\
    8th grade 2nd semester midterm & 0.0658 & \textbf{0.1787} & 0.1667\\
    8th grade 2nd semester final & -0.0092 & 0.0728 & \textbf{0.1372}\\
    \hline
    9th grade 1st semester midterm & 0.1557 & 0.1651 & \textbf{0.3410}\\
    9th grade 1st semester final & 0.1136 & 0.0460 & \textbf{0.2537}\\
    9th grade 2nd semester midterm & 0.2285 & 0.2334 & \textbf{0.3146}\\
    9th grade 2nd semester final & 0.0992 & 0.1236 & \textbf{0.1470}\\
    \bottomrule
    Average & 0.0972 & 0.1001 & \textbf{0.2130}\\
    \bottomrule
  \end{tabular}
  \caption{Average Knowledge level Improvement}
  \label{tab:knowledge-improve}
\end{table}

\begin{figure}[h]
  \centering
  \begin{subfigure}{0.33\textwidth}
    \centering
    \includegraphics[width=\linewidth]{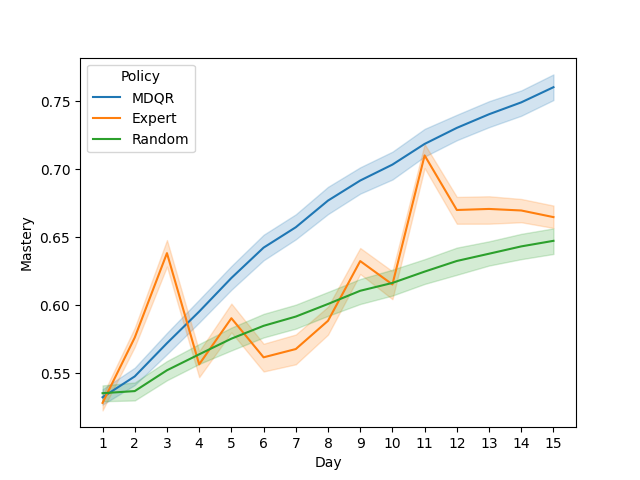}
    \caption{7th grade 1st semester midterm}
  \end{subfigure}%
  \begin{subfigure}{0.33\textwidth}
    \centering
    \includegraphics[width=\linewidth]{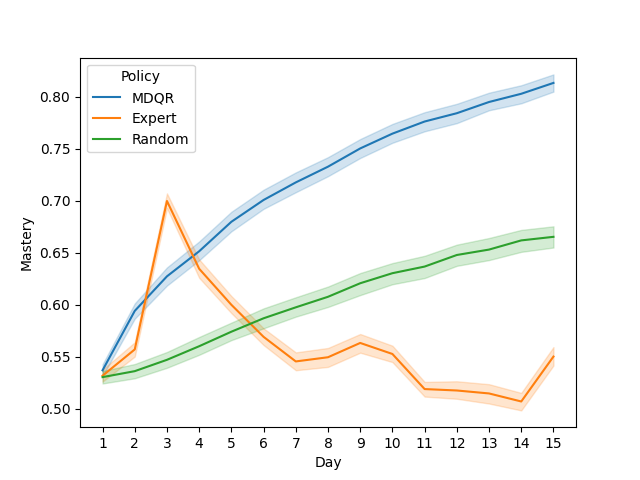}
    \caption{8th grade 1st semester midterm}
  \end{subfigure}%
  \begin{subfigure}{0.33\textwidth}
    \centering
    \includegraphics[width=\linewidth]{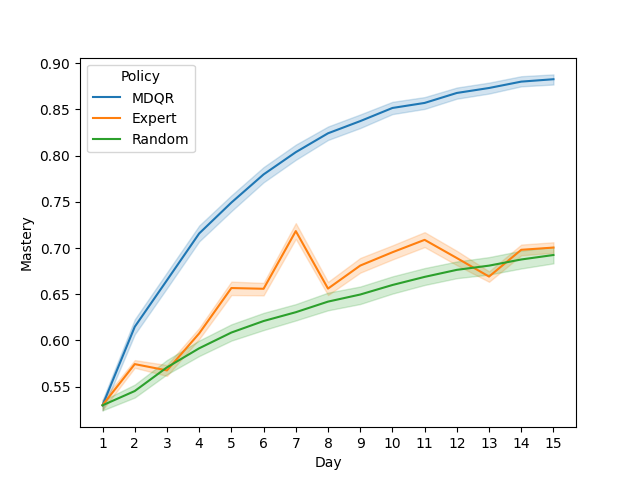}
    \caption{9th grade 1st semester midterm}
  \end{subfigure}
  \caption{Changes in knowledge level within the scope of each exam. Each curve shows the average across 2000 simulations under a Scheduling method. It also displays a 95\% confidence interval around the mean.}
  \label{fig:knowledge change}
\end{figure}

\subsection{Results}
As shown in Table \ref{tab:knowledge-improve}, in most test ranges, the MDQR increased the student's knowledge level the most during a given period. \textit{Expert} scheduler generally performed better than \textit{the random} scheduler but showed worse results in some test ranges. When looking at the average performance in each test range, MDQR showed \textbf{21.3\%} increase of knowledge level while Expert showed a 10\% increase. Figure \ref{fig:knowledge change} shows the change in the student's knowledge level over two weeks when the simulation was conducted with each question scheduling method. MDQR and the \textit{random} scheduler show a gradual rise in knowledge level, whereas the \textit{expert} scheduler shows unstable changes in knowledge level. MDQR and \textit{random} scheduler always recommend concepts of question in a different order, whereas \textit{expert} scheduler always recommends concepts of question in the same order according to the curriculum, with only different question difficulty. This shows that certain concepts cause a student's knowledge level to increase on average when solving questions on this concept while some concepts lower the knowledge level.

\section{Related Works}
There have been several attempts by many researchers to apply AI techniques to various fields in the educational domain \cite{chen2020aieducation}. Among them, the AI technique was mainly applied to the KT to analyze the knowledge level of the learner and the learning content scheduling technique for the efficient learning of the learner. In this section, we describe research in which AI technology is applied to KT and learning content scheduling.

\subsection{Knowledge Tracing}
KT is a technique to analyze and derive students' knowledge levels based on sequential learning log data. Corbett et al. \cite{corbett1994bkt} proposed Bayesian Knowledge Tracing (BKT), which analyzes the knowledge level using the Hidden Markov Model \cite{rabiner1986hmm} based on the correlation between the student's question-solving result and the knowledge level. Piech et al. \cite{piech2015dkt} proposed Deep Knowledge Tracing (DKT), which analyzes knowledge level based on LSTM, which is known to be proficient in dealing with sequential data. After DKT was proposed, deep learning-based KT techniques such as key-value memory network-based model (DKVMN) \cite{jiani2017dkvmn} and attention-based model (SAKT) \cite{pandey2019sakt} were proposed and showed high performance. In this paper, we utilized a deep learning-based KT model for a student question-solving simulator.

\subsection{Learning Contents Scheduling}
In addition to the KT technique, approaches to schedule appropriate learning content for students to learn efficiently have been widely studied. The key point of the learning content scheduling technique is to schedule the learning contents in the optimal order to maximize the evaluation measure such as the student's knowledge level or test score.

Ai et al. \cite{ai2019conceptaware} proposed a reinforcement learning-based approach to schedule exercises in an online learning system. Their method recommends the next question based on the current student's knowledge level. Specifically, they utilized reinforcement learning to derive an exercise that maximizes the reward calculated as the average of the student's knowledge level. They modeled the learning content scheduling process as a Partially Observable Markov Decision Process (POMDP) \cite{williams2007pomdp} and applied reinforcement learning based on the Trust Region Policy Optimization (TRPO) \cite{schulman2015trust} algorithm. In contrast, we applied a deep learning-based reinforcement learning model that autonomously learns content scheduling policies. Also, we solved the problem of recommending the same question repeatedly by introducing a scheduling penalty.

Loh et al. \cite{loh2021recommendation} suggested a question recommendation algorithm that can effectively raise a student's test score. They explained the phenomenon that recommendation algorithm based on the knowledge level of the student simply recommends only questions that a student is most likely to get right. To overcome this problem, they also considered the expected test scores assuming the student got the recommended questions right. They implemented a linear approximation-based student correct rate prediction model and a Bi-LSTM-based test score prediction model to predict students' test scores. 
Similarly, we introduce the concept of a penalty to solve the problem where the same question is recommended as duplicate. However, there are three differences from this study: First, we used the student's average knowledge level as a scheduling outcome measure instead of predictive test scores. Second, their method finds the next question with a greedy search based on the test score prediction model, whereas we find the optimal question sequence through a deep learning-based reinforcement learning model. Third, our study has the additional constraint that it should be scheduled among the questions that exist within the specific test range.

In addition, various course scheduling techniques to which artificial intelligence models are applied have been proposed. For example, a technique using reinforcement learning to determine the optimal order and number of activities provided within a 90-minute online class \cite{bassen2020reinforce}, and an approach to recommend a university course using catalog description and previous course enrollment history \cite{pardos2020serendipity}, etc. However, the purpose of our study is to schedule questions within a specific range, which is different from the corresponding studies.

\section{Conclusion}
This paper proposed MDQR, the first scheduling model that can adjust the recommendation range according to the student's test range. MDQR is a model that improved the DQN algorithm of reinforcement learning to fit our scheduling scenario. The purpose of the scheduling model is to maximize the learner's knowledge level within a limited period of time. In order to examine the performance of MDQR, we conducted a comparative evaluation between a \textit{random} scheduler and an \textit{expert} scheduler. Experimental results showed that MDQR outperforms in terms of learner's knowledge gain than other baselines in most of the test ranges.

\bibliographystyle{ACM-Reference-Format}
\bibliography{main}

\end{document}